\documentclass{IET-Conf-Paper}
\usepackage{tikz}
\usepackage{url}
\usetikzlibrary{arrows.meta,positioning,shapes.geometric}
\definecolor{TaskBlue}{RGB}{219,234,254}
\definecolor{AgentGreen}{RGB}{220,252,231}
\definecolor{OracleAmber}{RGB}{254,243,199}
\definecolor{FeedbackRose}{RGB}{255,228,230}
\definecolor{ProbeViolet}{RGB}{237,233,254}
\definecolor{DoneTeal}{RGB}{204,251,241}
\definecolor{EdgeGray}{RGB}{55,65,81}

\begin{document}

\title{COUNTEREXAMPLES AS FEEDBACK FOR AGENT SELF-CORRECTION}

\author{Sidhesh Badrinarayan\ad{1}, Adithya Parthasarathy\ad{1}}

\address{\add{1}{Senior IEEE Member}}

\keywords{multi-turn agents, agent evaluation, regular expression synthesis, LLM self-correction, LLM benchmarking}

\begin{abstract}
Single-turn code-generation metrics understate a central property of deployed agents: whether they can repair a wrong artifact after receiving concrete feedback. This paper presents A-CEGIS, a lightweight framework that uses counterexamples as feedback for evaluating multi-turn refinement in natural-language-to-regex synthesis. An agent proposes a regex, a deterministic oracle checks it under full-match semantics, and compact false-positive or false-negative witnesses guide the next turn. On 30 NL-RX-Turk tasks, diagnostic counterexample feedback solves 90\% of tasks within a four-turn ablation budget, compared with 17\% for zero-shot generation, 27\% for generic self-correction, and 23\% for error-only feedback. In a full diagnostic run with hardening, all tasks are solved on the hidden set by the final turn, with mean time-to-success of 2.7 turns and robust success of 77\% after targeted probing. These results show that A-CEGIS measures how efficiently an agent improves across turns while adding a practical robustness check beyond the original held-out cases.
\end{abstract}

\maketitle

\section{Introduction}
Software agents are rarely useful only on their first attempt: they must use feedback, repair wrong artifacts, and avoid breaking behaviour that was already correct. Standard one-shot metrics such as \textit{pass@1} remain useful, but they do not measure repair efficiency, stability across revisions, or whether an apparently solved task remains robust under harder tests.

We study this problem in natural-language-to-regex synthesis, where artifacts are compact, brittle, and executable. A small operator change can alter the accepted language, while correctness can be checked deterministically against positive and negative strings. This makes regex synthesis a controlled setting for isolating refinement behaviour without mixing it with retrieval, interface use, or other agent-environment effects.

This paper introduces A-CEGIS, a counterexample feedback loop for measuring agent self-correction. The agent proposes a regex, the oracle evaluates it with \texttt{re.fullmatch} semantics, and the next prompt receives concrete false-positive and false-negative witnesses rather than generic criticism. On 30 NL-RX-Turk tasks, diagnostic A-CEGIS reaches 0.900 hidden-set success within a four-turn ablation budget, compared with 0.167 for zero-shot generation, 0.267 for generic self-correction, and 0.233 for error-only feedback. With a longer diagnostic budget and targeted hardening, hidden-set success reaches 1.0 and robust success is 0.767. The contributions are a reproducible CEGIS-style loop, trajectory metrics for convergence and regressions, direct feedback-strategy ablations, and a robustness probe that exposes failures hidden by endpoint accuracy.

\section{Background and Related Work}
LLM evaluation has historically emphasised first-attempt quality, exact match, and single-step task completion. Code-generation benchmarks such as HumanEval and MBPP established functional correctness as a central measure for synthesis \cite{b8,b9}, while SWE-bench moves evaluation toward realistic software repair \cite{b12}. Recent work argues that multi-turn agents also require measures of context retention, interaction quality, and error recovery \cite{b1}; coding benchmarks similarly show that performance can degrade in multi-turn settings \cite{b2}. These results motivate process-level evaluation rather than endpoint scoring alone.

Self-refinement methods show that models can improve outputs through repeated feedback and revision \cite{b3,b11}, and ReAct demonstrates the value of interleaving reasoning with environment actions \cite{b10}. However, natural-language feedback can be noisy or underspecified. AgentBoard and related benchmarks broaden evaluation to tool-mediated environments \cite{b4}, but such settings can mix repair skill with retrieval, action selection, and interface effects. A-CEGIS asks a narrower complementary question: given deterministic evidence of failure, can an agent repair a compact formal artifact efficiently?

Natural-language-to-regex synthesis fits this question well. NL-RX maps descriptions to regular expressions \cite{b5}, while programming-by-example and CEGIS work show the value of executable examples and counterexample-guided revision \cite{b14,b6,b13}. A-CEGIS adapts that interaction pattern to language-agent evaluation. It is not a full equivalence proof, since the verifier uses finite tests and targeted probes, but every turn is grounded in executable semantics. This lets us measure how many turns are needed, whether repairs are local, whether revisions introduce regressions, and whether a hidden-set solution survives additional probing.

\section{Methodology}
The evaluation harness uses deterministic Python regex evaluation, synthetic task-local tests, and an LLM interface. A task is $(d_i,g_i)$, where $d_i$ is a natural-language description and $g_i$ is a gold Python-compatible regex. For each task, the oracle builds
\[
T_i=P_i\cup N_i,
\]
with 12 generated positives and 12 generated negatives in the reported configuration. Positives are sampled from the gold regex using a lightweight parser-driven generator; negatives come from mutations and random samples retained only when rejected by $g_i$.

\subsection{Oracle construction}
The gold regex is used only as an executable specification. Positive examples are generated by recursively traversing the parsed expression when possible, including literals, branches, character classes, subpatterns, and bounded repetitions. Negative examples are generated from positive seeds by deletion, insertion, substitution, swapping, truncation, and random sampling over a printable alphabet, following the intuition that small edit operations often expose boundary behaviour \cite{b7}. A sampled negative is retained only if it fails the gold regex under full-string matching.

This construction has two practical consequences. First, the benchmark can be run without an external theorem prover or automata library. Secondly, the hidden set is task-local: failures tend to occur near the semantic material of the target regex rather than in an unrelated global string distribution. The trade-off is that hidden-set success is not semantic equivalence. That limitation is deliberate, because it creates room to measure the difference between sampled success and robustness under targeted probing.

\begin{figure}[h]
\begin{center}
\resizebox{\linewidth}{!}{\begin{tikzpicture}[
  node distance=7mm,
  block/.style={draw=EdgeGray, rounded corners, align=center, minimum width=25mm, minimum height=7mm, font=\scriptsize, line width=0.45pt},
  decision/.style={draw=EdgeGray, diamond, aspect=2.0, align=center, inner sep=1pt, font=\scriptsize, line width=0.45pt},
  arr/.style={-Latex, thick, draw=EdgeGray}
]
\node[block, fill=TaskBlue] (task) {Task\\description $d_i$};
\node[block, fill=AgentGreen, right=of task] (agent) {Agent\\proposes $r_t$};
\node[block, fill=OracleAmber, right=of agent] (oracle) {Oracle\\checks $T_i$};
\node[decision, fill=OracleAmber, below=of oracle] (pass) {all\\pass?};
\node[block, fill=FeedbackRose, below=of agent] (feedback) {False pos./neg.\\counterexamples};
\node[block, fill=ProbeViolet, below=of pass] (harden) {Targeted\\hardening probes};
\node[decision, fill=ProbeViolet, below=of harden] (clean) {probe\\clean?};
\node[block, fill=DoneTeal, right=of clean] (done) {Robust\\solution};
\draw[arr] (task) -- (agent);
\draw[arr] (agent) -- (oracle);
\draw[arr] (oracle) -- (pass);
\draw[arr] (pass) -- node[above,font=\scriptsize]{no} (feedback);
\draw[arr] (feedback.north) -- (agent.south);
\draw[arr] (pass) -- node[right,font=\scriptsize]{yes} (harden);
\draw[arr] (harden) -- (clean);
\draw[arr] (clean) -- node[above,font=\scriptsize]{yes} (done);
\draw[arr] (clean.west) -| node[pos=.25,above,font=\scriptsize]{no} (feedback.south);
\end{tikzpicture}}
\caption{A-CEGIS evaluation loop in which candidate regexes are checked against hidden tests, failures become diagnostic counterexamples, and hidden-set success triggers targeted robustness probing.}
\label{fig:acegis}
\end{center}\vspace*{-18pt}
\end{figure}
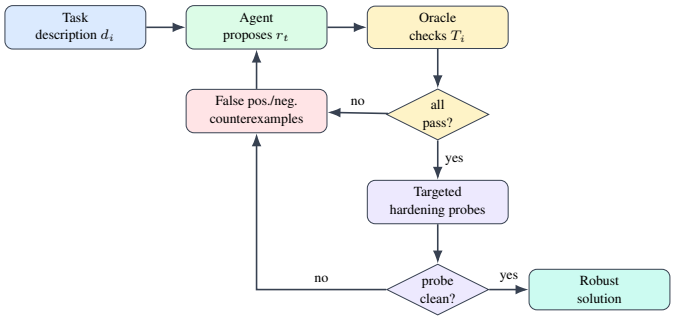

Figure~\ref{fig:acegis} shows the core loop. At turn $t$, the agent proposes $a_i^{(t)}$. The oracle computes the pass vector $\mathbf{y}_i^{(t)}$ over $m=|T_i|$ cases and pass rate
\[
\rho_i^{(t)}=\frac{1}{m}\sum_{j=1}^{m}y_{i,j}^{(t)}.
\]
If $\rho_i^{(t)}<1$, feedback is produced according to the strategy. In the diagnostic A-CEGIS strategy, feedback contains up to two false negatives and two false positives, sorted to favour compact witnesses. This tells the model not only that the regex is wrong but also whether the language should be broadened or tightened.

The implementation supports a small strategy grid: zero-shot generation, self-correction with generic language feedback, error-only feedback with no witnesses, minimal A-CEGIS with one failing example, three-example A-CEGIS, and diagnostic A-CEGIS. The ablation reported below compares zero-shot, generic self-correction, error-only feedback, and diagnostic A-CEGIS on the same task set. The common loop is the same across strategies; only the information returned after a failed turn changes.

After hidden-set success, hardening searches a targeted probe pool $Q_i(r)$ for witnesses where $g_i$ and $r$ disagree. The pool is built from short generic strings, tokens in the task description, literals extracted from the gold and candidate regexes, existing positive and negative examples, and simple transformations such as prefixing, suffixing, repetition, concatenation, and character mutation. This makes the probe distribution intentionally local: it stresses boundaries near the language already implied by the task rather than sampling arbitrary strings.

\subsection{Hardening protocol}
Hardening is intended to expose semantic errors that the original hidden set did not sample. For a candidate $r$, the probe stage searches for strings
\[
w\in Q_i(r)\quad\text{such that}\quad \mathbb{1}[g_i(w)]\neq \mathbb{1}[r(w)].
\]
Each mismatch is labelled by direction. If the gold regex accepts $w$ but the candidate rejects it, $w$ becomes a probe-derived false negative; if the candidate accepts $w$ but the gold rejects it, $w$ becomes a probe-derived false positive. In the reported configuration, each hardening cycle evaluates up to 320 probes, appends at most four probe-derived counterexamples to the task-local test set, and gives the agent one repair attempt. This process is repeated for at most two cycles, producing an expanded set $T_i'=T_i\cup C_i$.

The final status is stricter than base success. \emph{Hidden success} means that a candidate passes the original sampled positives and negatives. \emph{Hardened success} means that the repaired candidate passes the expanded set after probe-derived examples are added. \emph{Robust success} additionally requires a final probe pass with no remaining mismatches. Thus a regex can appear solved on the hidden set and still fail robustness because it accepts decorated literals, misses minimum-length cases, or uses an overly broad character class.

The benchmark records endpoint and process metrics. If $\tau_i$ is the first successful turn, it reports Pass@1, Pass@final, mean time-to-success, and MRR. Repair locality is measured by normalised edit similarity between consecutive regex token sequences,
\[
S_i^{(t)}=1-\frac{\mathrm{EditDist}(u^{(t-1)},u^{(t)})}{\max\{|u^{(t-1)}|,|u^{(t)}|,1\}},
\]
and net repair behaviour is measured by the refinement efficiency ratio
\[
\mathrm{RER}=\frac{\sum_{i,t} f_i^{(t)}}{\max(1,\sum_{i,t} b_i^{(t)})},
\]
where $f$ counts fixed cases and $b$ counts regressions.

These metrics are designed to describe the refinement trajectory rather than only its endpoint. Mean time-to-success and MRR capture how quickly the agent converges. Stability distinguishes local repair from complete regeneration. RER asks whether the loop fixes more behaviours than it breaks, which is especially important for formal artifacts: a revision that solves one counterexample while regressing several earlier cases is not good repair, even if it looks plausible in natural language.

\section{Experiments and Results}
The experimental artifacts evaluate Gemini 3 Flash Preview on 30 NL-RX-Turk tasks. Each task uses 24 hidden oracle checks before hardening. We report two complementary settings. First, a four-turn feedback-strategy ablation compares zero-shot generation, generic self-correction, error-only feedback, and diagnostic A-CEGIS without hardening. Secondly, a full diagnostic run allows up to seven turns, followed by at most two hardening cycles with one repair attempt per cycle.

The run is intentionally focused on feedback mechanisms rather than model ranking. It does not claim to compare model families. Instead, it tests whether a capable model, given structured counterexamples, exhibits the repair dynamics that A-CEGIS is meant to measure and whether those dynamics differ from weaker feedback channels. Because regex evaluation is deterministic, changes in pass rate, stability, and robustness can be attributed to the candidate sequence rather than to stochastic environment effects.

\begin{table}[h]
\caption{Full diagnostic A-CEGIS results over 30 tasks}
\begin{center}
\begin{tabular}{ll}\toprule
Metric & Value \\\midrule
Tasks & 30 \\
Pass@1 & 0.167 \\
Pass@final (hidden set) & 1.000 \\
Mean turns-to-success & 2.667 \\
Mean reciprocal rank & 0.477 \\
Mean initial pass rate & 0.706 \\
Mean final pass rate & 1.000 \\
Mean structural stability & 0.561 \\
Fix events & 311 \\
Break events & 99 \\
Refinement efficiency ratio & 2.613 \\
Hardened pass@final & 0.767 \\
Probe-clean rate & 0.767 \\
Robust pass@final & 0.767 \\
Mean hardened final pass rate & 0.937 \\
Probe-derived counterexamples added & 80 \\
Hardening cycles completed & 40 \\\botrule
\end{tabular}
\label{tab:mainresults}
\end{center}\vspace*{-18pt}
\end{table}

Table~\ref{tab:mainresults} shows a large gap between first-turn and final hidden-set success. Only 5 of 30 tasks are solved immediately, yet all are solved by the full diagnostic loop. The mean initial pass rate of 0.706 indicates that initial regexes are often close but incomplete; the repair trajectory turns those partial matches into hidden-set-perfect solutions. RER is 2.61, with 311 fixes against 99 breaks, so refinement is net constructive rather than oscillatory.

The MRR of 0.477 shows that the gain is not merely a final-turn rescue after many attempts. Most tasks converge within a few turns, but not on the first turn. This is exactly the regime in which multi-turn evaluation is informative: one-shot accuracy is too pessimistic about eventual repair ability, while final success alone hides the cost and path of convergence.

\begin{table}[h]
\caption{Feedback-strategy ablation over 30 tasks with a four-turn budget}
\begin{center}
\scriptsize
\setlength{\tabcolsep}{2pt}
\begin{tabular}{lrrrr}\toprule
Strategy & Pass@final & MRR & RER & Attempts \\\midrule
Zero-shot & 0.167 & 0.167 & 0.000 & 1.00 \\
Self-correction & 0.267 & 0.211 & 0.674 & 2.77 \\
Error-only & 0.233 & 0.194 & 0.990 & 2.83 \\
Diag. A-CEGIS & 0.900 & 0.458 & 2.521 & 2.53 \\\botrule
\end{tabular}
\label{tab:ablation}
\end{center}\vspace*{-18pt}
\end{table}

Table~\ref{tab:ablation} isolates the effect of feedback content. All strategies use the same model and task set, and the diagnostic row is evaluated under the same four-turn budget as the baselines. Returning concrete false-positive and false-negative witnesses is substantially more effective than asking the model to self-correct generically or reporting only the number of hidden-test failures. Diagnostic A-CEGIS solves 27 of 30 tasks within four turns, compared with 8 for generic self-correction, 7 for error-only feedback, and 5 for zero-shot generation. It also uses no more average generation attempts than the weaker multi-turn baselines, indicating that the improvement comes from more informative feedback rather than simply more turns.

\begin{table}[h]
\caption{Distribution of first successful turn}
\begin{center}
\begin{tabular}{ll}\toprule
Success turn & Number of tasks \\\midrule
1 & 5 \\
2 & 10 \\
3 & 9 \\
4 & 3 \\
5 & 2 \\
6 & 1 \\\botrule
\end{tabular}
\label{tab:turns}
\end{center}\vspace*{-18pt}
\end{table}

Table~\ref{tab:turns} shows that most tasks require explicit repair: 25 of 30 are solved after turn 1, and the mode is turn 2. No task requires more than six turns. The hardening results then separate hidden-set success from robustness. Only 23 of 30 tasks remain probe clean; the other seven pass the original sampled tests but fail on targeted variants involving boundaries, repetitions, or broad character classes.

The turn distribution also suggests that diagnostic counterexamples narrow the search space quickly. The system is not repeatedly sampling unrelated regexes until one happens to pass. Instead, many trajectories make a small number of semantically meaningful changes, such as replacing \texttt{.*} with \texttt{.+}, tightening an alternation, or adding a missing character-class constraint.

\begin{table}[h]
\caption{Representative hardening examples}
\begin{center}
\scriptsize
\setlength{\tabcolsep}{3pt}
\begin{tabular}{@{}p{3.4cm}p{4.6cm}@{}}\toprule
Task description & Repair \\\midrule
Letter preceded by a number (turn 4) &
\textit{Base:} \texttt{[\string^a-zA-Z]*[0-9].*[a-zA-Z].*}\newline
\textit{Hardened:} \texttt{[0-9].*} \\\midrule
Lower-case letter, letter, or character (turn 2) &
\textit{Base:} \texttt{.}\newline
\textit{Hardened:} \texttt{\textbackslash w} \\\midrule
Contains `dog', a character, or a numeral (turn 2) &
\textit{Base:} \texttt{.*dog.*\textbar\textbackslash S}\newline
\textit{Hardened:} \texttt{dog\textbar[A-Za-z0-9\_]} \\\midrule
Lower-case letter or letter, seven or more times (turn 2) &
\textit{Base:} \texttt{.*[a-zA-Z]\{7,\}.*}\newline
\textit{Hardened:} \texttt{[a-zA-Z]\{7,\}} \\\botrule
\end{tabular}
\label{tab:examples}
\end{center}\vspace*{-18pt}
\end{table}

Table~\ref{tab:examples} illustrates the common pattern. The model often captures the coarse language but expresses it with permissive constructs such as \texttt{.*}, broad one-character classes, or unnecessary context around a literal. Hardening can remove this slack: \texttt{.*dog.*\textbar\textbackslash S} is tightened to \texttt{dog\textbar[A-Za-z0-9\_]}, and surrounding wildcards are removed from an expression requiring seven or more letters. These cases explain both the 0.561 stability score and the robustness gap: repairs preserve broad structure while tightening operators that leak semantics.

\subsection{Failure modes}
The seven non-robust cases fall into three recurring categories. The first is boundary under-specification, where a regex accidentally accepts the empty string, a one-character string, or a missing suffix/prefix case. The second is wildcard overreach, where \texttt{.*} admits repeated or decorated strings that were absent from the base hidden set. The third is class-boundary confusion, where a broad class such as \texttt{\textbackslash w} or a negated class includes more symbols than the description intended.

These failures are not merely anecdotal edge cases. They show that hidden-set success can validate a sampled behaviour profile while still missing the intended language shape. In that sense, the hardening phase plays a different role from ordinary testing: it is not just another larger test set, but a targeted search around likely semantic boundaries.

\subsection{Interpretation}
The main empirical lesson is that the same run can look excellent or incomplete depending on the measurement layer. Under hidden tests, the method solves every task. Under trajectory metrics, it shows efficient and mostly constructive repair. Under hardening, it reveals that roughly one quarter of hidden-set-perfect solutions still contain latent semantic errors. These statements are not contradictory; they describe different levels of reliability.

For agent evaluation, this distinction is useful. A benchmark that reports only Pass@1 would miss the model's ability to recover. A benchmark that reports only final hidden success would miss the residual robustness gap. A-CEGIS makes both visible in one compact loop.

\section{Example A-CEGIS Run}
To make the protocol concrete, consider a typical task trajectory. The model often begins with a regex that captures the main lexical cue but misses a boundary condition. The oracle evaluates the candidate on all positives and negatives, separates failures by direction, and returns a few witnesses. A false negative says that the expression is too narrow; a false positive says that it is too broad. The next candidate is therefore conditioned on a behavioural correction rather than a vague request for improvement.

The traces make the run auditable. A close first candidate may need only a local edit, such as replacing \texttt{.*} with \texttt{.+} after a minimum-length failure, while a structurally wrong candidate may require a larger rewrite. Structural stability and RER distinguish these cases by recording whether edits are local and whether they fix more cases than they break. The released artifacts include aggregate metrics, per-task candidate sequences, and probe-derived counterexamples, so a run can be inspected rather than only scored. The code is available in the public A-CEGIS repository \cite{b16}.

\section{Discussion}
A-CEGIS is best understood as a measurement protocol for agent repair. Its value is that it separates refinement behaviour from other benchmark complexity: the artifact is compact, the verifier is deterministic, and feedback is concrete behavioural evidence. The ablation shows why this matters. Diagnostic A-CEGIS reaches 0.900 Pass@final within four turns while generic self-correction and error-only feedback remain below 0.300, indicating that directional witnesses help the model make targeted edits instead of inferring the failure mode from a scalar score.

The trajectory metrics add information that endpoint scores cannot provide. Pass@1 shows the difficulty of initial synthesis, final hidden success shows convergence, RER records whether revisions are constructive, and stability distinguishes local repairs from wholesale rewrites. Targeted hardening then adds an active robustness layer after hidden-set success by probing strings near description tokens, regex literals, and existing examples. This checks whether a recovered expression remains reliable near the semantic boundaries implied by the task.

A-CEGIS is meant to generalise beyond regexes wherever three ingredients are available: a compact formal artifact, a deterministic oracle, and counterexamples that can be rendered back to the model. SQL queries can be checked against databases, schemas against documents, and configuration policies against generated states. Recent DocSync work applies a related critic-guided refinement loop to documentation maintenance, pairing structural code context with iterative critique to keep descriptions aligned with implementation \cite{b15}. In each case, the core question is whether an agent can use concrete evidence to make a targeted repair.

\section{Conclusion and Future Work}
A-CEGIS shows that counterexamples are a powerful feedback signal for agent self-correction. Instead of scoring only the first answer or final endpoint, the framework records the proposed artifact, concrete behavioural failures, revisions, regressions, and robustness under additional probing. In regex synthesis, diagnostic counterexamples raise four-turn Pass@final to 0.900 and full diagnostic hidden-set success to 1.0, while hardening exposes a remaining robustness gap. Future work can scale the benchmark across additional models, add exact equivalence checks where available, and extend the same compact-artifact, deterministic-oracle pattern to SQL queries, schemas, configuration policies, and program fragments.

The main lesson is that refinement quality should be evaluated as a trajectory, not as a single number. A final answer can pass sampled tests while still being brittle, and a low first-attempt score can hide a strong ability to repair once concrete evidence is available. By preserving the sequence of candidates, counterexamples, fixes, and breaks, A-CEGIS makes those distinctions visible and gives future evaluations a practical way to compare not only whether agents succeed, but how they improve.

\end{document}